\documentclass[10pt,twocolumn,letterpaper]{article}

\usepackage{iccv}
\usepackage{times}
\usepackage{epsfig}
\usepackage{graphicx}
\usepackage{amsmath}
\usepackage{amssymb}
\usepackage{ctable} 
\usepackage{mathrsfs}
\usepackage[symbol*]{footmisc}
\usepackage{multirow}
\usepackage{tablefootnote}
\usepackage{float}
\usepackage{color}
\usepackage{subcaption}
\usepackage[pagebackref=true,breaklinks=true,letterpaper=true,colorlinks,bookmarks=false]{hyperref}

\newcommand*{\affaddr}[1]{#1} 
\newcommand*{\affmark}[1][*]{\textsuperscript{#1}}
\newcommand*{\email}[1]{\texttt{\small#1}}

\iccvfinalcopy 

\def\iccvPaperID{12} 

\begin{document}
\title{Self-Supervised Pretraining and Controlled Augmentation Improve Rare Wildlife Recognition in UAV Images}


\author{
Xiaochen Zheng\affmark[1,2]\quad Benjamin Kellenberger\affmark[1]\quad Rui Gong\affmark[3]\quad Irena Hajnsek\affmark[2,4]\quad Devis Tuia\affmark[1]\\
\affaddr{\affmark[1]ECEO, EPFL\quad \affmark[2]IfU, ETH Zürich\quad \affmark[3]CVL, ETH Zürich\quad \affmark[4]DLR}\\
\email{xzheng@student.ethz.ch\quad gongr@vision.ee.ethz.ch\quad irena.hajnsek@dlr.de}\\
\email{\{benjamin.kellenberger, devis.tuia\}@epfl.ch} \\
}

\maketitle

\begin{abstract}
Automated animal censuses with aerial imagery are a vital ingredient towards wildlife conservation. Recent models are generally based on deep learning and thus require vast amounts of training data. Due to their scarcity and minuscule size, annotating animals in aerial imagery is a highly tedious process. In this project, we present a methodology to reduce the amount of required training data by resorting to self-supervised pretraining. In detail, we examine a combination of recent contrastive learning methodologies like Momentum Contrast (MoCo) and Cross-Level Instance-Group Discrimination (CLD) to condition our model on the aerial images without the requirement for labels. We show that a combination of MoCo, CLD, and geometric augmentations outperforms conventional models pretrained on ImageNet by a large margin. Crucially, our method still yields favorable results even if we reduce the number of training animals to just 10\%, at which point our best model scores double the recall of the baseline at similar precision. This effectively allows reducing the number of required annotations to a fraction while still being able to train high-accuracy models in such highly challenging settings.




\end{abstract}

\section{Introduction}
\label{intro}

\begin{figure}[t]
\begin{center}
  \includegraphics[width=\linewidth]{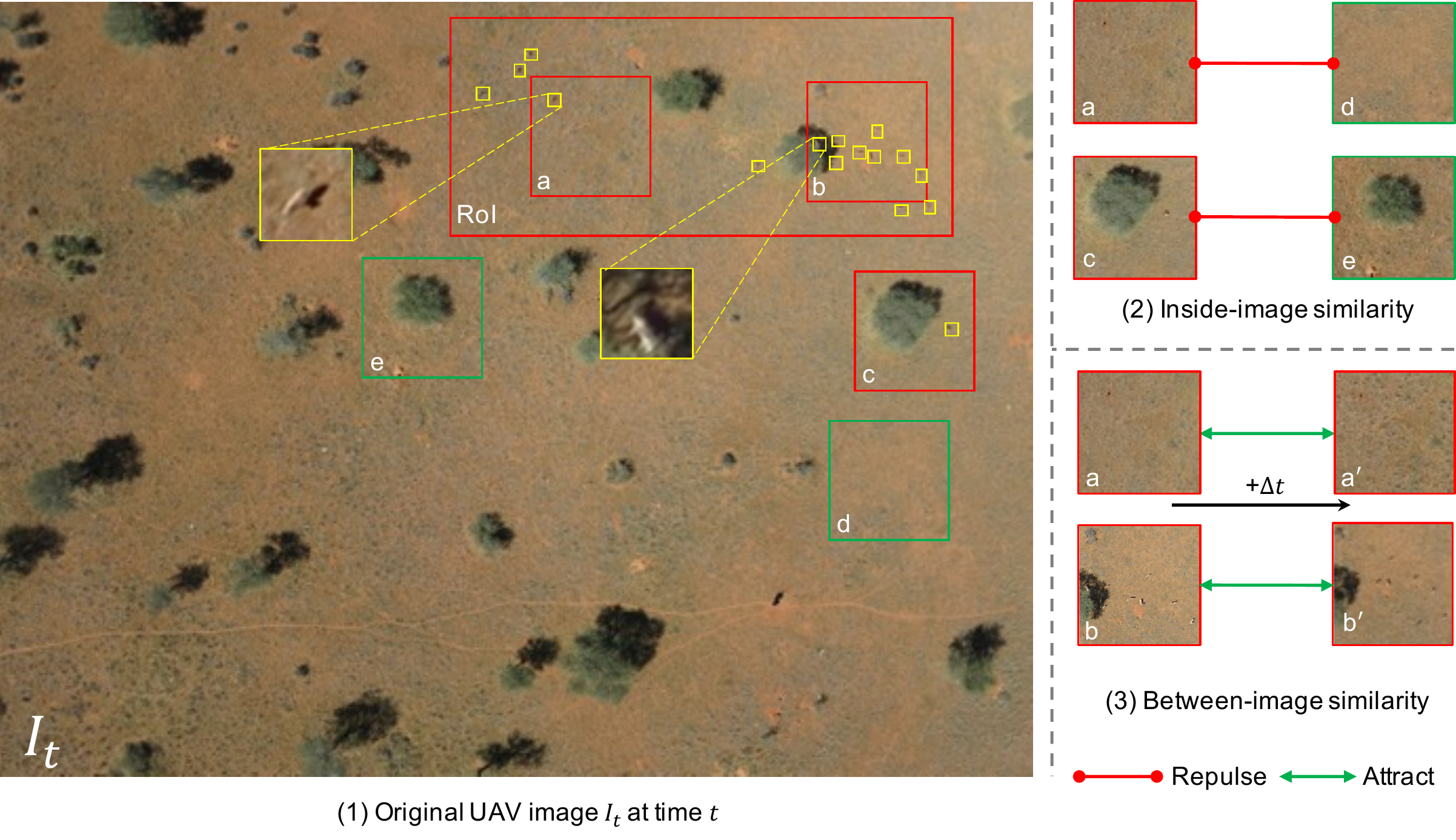}
\end{center}
\caption{Overview of Kuzikus dataset. In (1), the region of interest (RoI) (\textcolor[RGB]{255,0,0}{red} bounding box) only occupies a small part of original UAV image. Wildlife (\textcolor{yellow}{yellow} bounding boxes) is tiny which is hard to be recognized and labeled. That makes supervised learning difficult. Foreground (wildlife)/background crops are marked with \textcolor{red}{red}/\textcolor{green}{green} bounding boxes. Patches $a, c, d, e$ are cropped from different locations of the same image. Patches $a, a^\prime, b, b^\prime$ are cropped from same locations of different images (sampling time interval is $\Delta t$). In (2) similar patches containing wildlife should be distinguished (repulsed) from the ones without wildlife. Whereas in (3) similar patches from the same category should be attracted. (2) and (3) are difficult for supervised learning and contrastive learning.}
\label{fig:correlation}
\end{figure}

Wildlife censuses help determine the exact number and the spatial-temporal distribution of wild animals, which is vital to assess living conditions and potential survival risks of wildlife species~\cite{Beni_2018_RSE,balmford2005convention,biodiversity_review}. Since recently, these hitherto manual surveys of wildlife reserves are increasingly replaced with counts derived automatically from images acquired by Unmanned Aerial Vehicles (UAVs), paired with deep learning models to automate the wildlife recognition task~\cite{Beni_2018_RSE, Beni_al, Devis_detecting, Kellenberger_2019_CVPR_Workshops,eikelboom2019improving}.
These models are typically supervised, pretrained on large-scale curated datasets such as ImageNet~\cite{imagenet_cvpr09}, MS-COCO~\cite{coco} and then fine-tuned on the target imagery. Irrespective of the type of fine-tuning, this last step requires thousands of animals to be annotated, and hence expert knowledge. Meanwhile, large UAV campaigns can generate images in high numbers~\cite{Beni_2018_RSE}, with many containing either large numbers of animals, or none~\cite{Kellenberger_2019_CVPR_Workshops}, both of which cases are cumbersome for manual annotators. Furthermore, the vastness of wildlife reserves mean that wildlife is a rare sight and background dominates the majority of images. This causes two problems: on the one hand, the datasets are strongly imbalanced toward the absence class; on the other hand, the objects of interest are extremely small compared with the original aerial image size, as illustrated in Figure~\ref{fig:correlation}. These two issues significantly downgrade the capacity of a supervised model to solve the recognition task~\cite{Beni_2018_RSE}. Hence, methods to reduce the labelling requirement are urgently needed.

A promising direction to this end is to use self-supervised learning (SSL), where models are first trained in a \emph{pretext} task on the target images without the need for manually provided labels, and then fine-tuned on the actual objective (\emph{downstream task}) with manual labels~\cite{invariant}. Earlier \emph{pretext} tasks required models to reconstruct transformations between different \emph{views} of the same image. Recently, focus has shifted to contrastive learning. Here, augmentation as a method of image transformation is still employed, but with a different objective: while traditional SSL methodologies forced the model to learn representations within one data point to \emph{e.g.}, in-paint cut-out regions~\cite{inpainting, zhang2016colorful}, contrastive learning employs transformations in a comparison scheme and encourages the model to learn representations by maximizing the similarity between two randomly augmented \emph{views} of the same data point, resp. image (positive pairs) and dissimilarity between different data points (negative pairs)~\cite{Dahua_2018_CVPR, he2020momentum, mocov2, cmc, SimCLR, byol, PIRL_2020_CVPR}.
The different \emph{views} of same instance are randomly generated from a stochastic data augmentation module. Recent works have identified a stronger augmentation strategy to be vital for improved learning~\cite{mocov2}. However, augmentation functions need to be carefully selected with respect to the problem and data at hand~\cite{LooC}. Choosing inadequate functions may result in removal of important information (\emph{e.g.}, random resized cropping may remove animals at the border of UAV images). In contrast, some functions benefit certain scenarios more than others (\emph{e.g.}, random vertical flips and rotations may be of limited use with natural images, but may provide strong learning signals for view-independent aerial images).

However, applying self-supervised learning techniques, such as contrastive learning, to UAV images on wildlife is challenging.
One such problem is the requirement of contrastive learning methods to receive dissimilar imagery. In UAV acquisitions, as illustrated in Figure~\ref{fig:correlation}, high image sampling frequencies will generate strong autocorrelations between acquisitions in short time intervals (similar to adjacent frames in a video). Also, the vastness of many wildlife areas result in consistent characteristics, and hence in repeated or similar patches in the dataset. Many instance discrimination based methods, such as NPID~\cite{Dahua_2018_CVPR}, MoCo~\cite{he2020momentum, mocov2}, and SimCLR~\cite{SimCLR}, are based on the assumption that each instance is significantly different from others and that each instance can be treated as a separate category. The large similarity between training images mean that the negative pairs used in the contrastive learning process is likely to be composed of highly similar instances, which in turn compromises feature representations due to incorrect repulsion between similar images. Instead of exploring the effect of hard negative sampling~\cite{hard_neg_mix,hard_neg_sample}, in this paper we propose to solve these problems with Cross-Level Instance-Group Discrimination (CLD)~\cite{Wang_2021_CVPR}, which aims to deal with highly correlated datasets. Furthermore, we hypothesize that top view UAV imagery should be \emph{invariant} to geometric transformations, \emph{e.g.}, rotation. From this perspective, we propose to apply extra geometric transformation to contrastive model, which captures invariant information of UAV images introduced by different augmentations. 


We propose an SSL model to pretrain wildlife recognition models based on contrastive learning. Our work build on the work of MoCo~\cite{he2020momentum,mocov2} and Cross-Level Instance-Group Discrimination (CLD)~\cite{Wang_2021_CVPR}:
\begin{itemize}
    \item We propose a methodology for image-level wildlife recognition with a reduced number of annotations by self-supervised pretraining.
    \item We show that using self-supervised pretraining outperforms supervised ImageNet pretraining on downstream recognition task.
\end{itemize}
We further find that applying controlled augmentation to self-supervised pretraining and fine-tuning the pretrained model with few labels will outperform ImageNet pretraining fine-tuned with all available training labels. Our self-supervised pretraining learns representations of natural wildlife scenes more efficiently than supervised pretraining.

\section{Related Work}


\begin{figure*}[t]
\begin{center}
  \includegraphics[width=\linewidth]{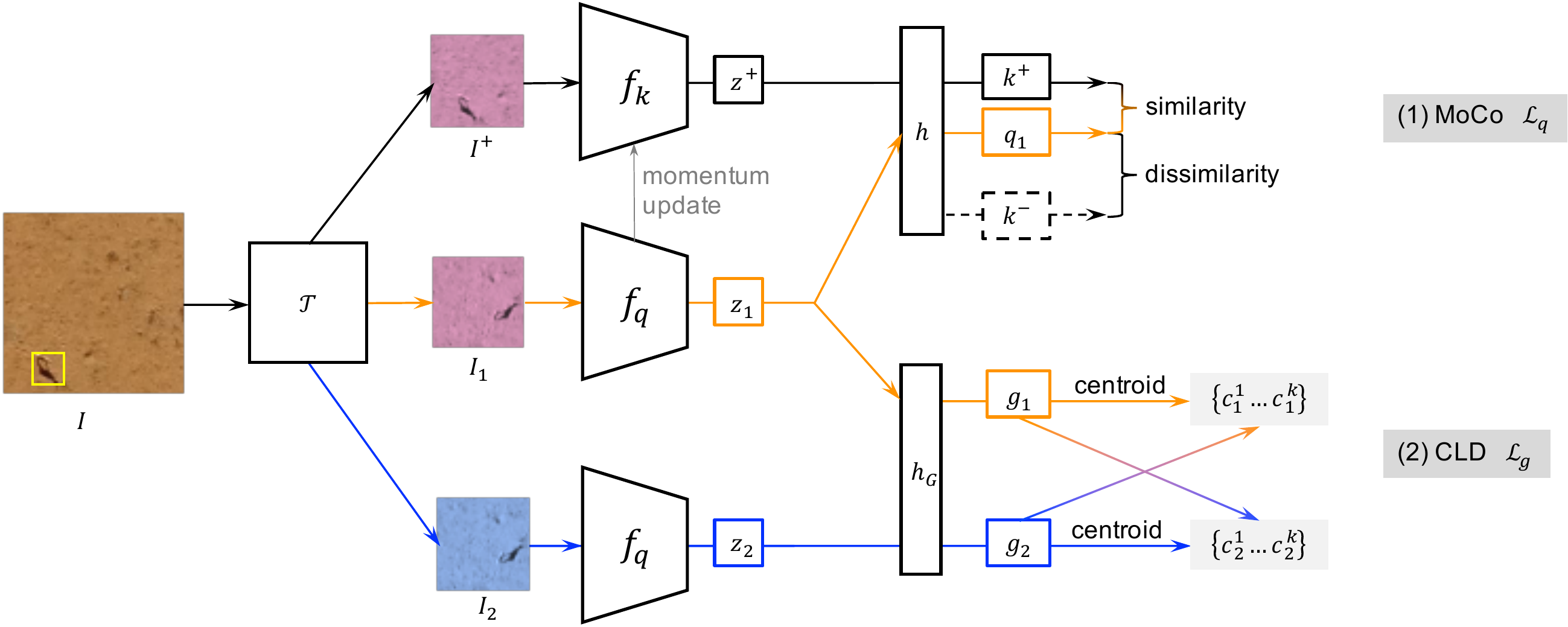}
\end{center}
  \caption{Overview of the employed SSL framework, consisting of MoCo~\cite{he2020momentum} (upper part) and CLD~\cite{Wang_2021_CVPR} (lower part). Firstly, the two different \emph{views} $I_1$ and $I^+$ of the same input $I$ are encoded by $f_q$ and $f_k$ respectively. Then, the two representations $z_1$ and $z^+$ are projected into an embedding space. $q$ and $k$ are representations of query and key in the hypersphere. Query $q_1$ and its positive key $k^+$ are from the different augmented \emph{views} of the same input and negative keys $k^-$ are encoded from different inputs (dashed bounding box). CLD first encodes two different \emph{views} $I_1, I_2$ of the same instance, then applies a different projection head and projects the representations from the same query encoders to a different embedding space from (1). Finally, a local K-Means clustering is used to find the $k$ centroids of a batch of inputs. The centroid of assigned cluster $g_1$ can be served as positive key of view $g_2$, and vice versa.}
\label{fig:model}
\end{figure*}

\textbf{Self-Supervised Learning in UAV imagery.}
Unlike in the field of classic computer vision, self-supervised learning of aerial images has not yet been fully studied. Stojnic~\etal~\cite{RMSSL_Stojnic_2021_CVPR} apply Contrastive Multiview Coding~\cite{tian2020view} to learn aerial image representations on both RGB and multispectral remote sensing images. \cite{kang2020deep} proposes a method based on contrastive learning with different image augmentations. \cite{tao2020remote} analyze different pretext tasks, \emph{e.g.}, inpainting~\cite{inpainting}, context prediction~\cite{context_prediction}, and contrastive learning with different image augmentations on remote sensing dataset. Besides, \cite{ayush2020geography} use geo-location classification as the pretext task. The encoder is trained by predicting the global geo-location of input image. Tong~\etal~\cite{land-cover} generate pseudo-labels of unlabeled UAV imagry data to improve the downstream classification accuracy. Most of the downstream tasks in aerial imaery domain are scene classification, \emph{e.g.}, land-cover or land-use~\cite{jean2019tile2vec} classification. \cite{ayush2020geography}  apply self-supervised learning to transferred downstream tasks, \emph{e.g.}, object detection and image segmentation. Different from those tasks, our task is more domain specific, which is tiny and rare wildlife recognition in the wild.

\textbf{Pretext tasks.}
Self-supervised representation learning is designed to solve certain pretext tasks. \cite{rotation} uses the rotation angle as pseudo label and learn underlined structure of the objects by predicting rotation angle. \cite{context_prediction, invariant, ding2021unsupervised} perform region-level relative location prediction. Other missions like color in-painting~\cite{inpainting, zhang2016colorful}, and solving jigsaw puzzles~\cite{jigsaw} are also applied as pretext tasks. In contrastive learning, learning between different augmented views are used as pretext tasks.  In this work, we apply a instance discrimination task~\cite{Dahua_2018_CVPR} in addition to geometric invariant mapping to contrastive self-supervised models.

\textbf{Contrastive Representation Learning}.
Recently, the most competitive representation learning method without labels is self-supervised contrastive learning. Contrastive methods~\cite{SimCLR,he2020momentum,mocov2,byol,SwAV,cpc,invariant,lecun_invariant_mapping, PIRL_2020_CVPR} train a visual representation encoder by attracting positive pairs from the same instance in latent space while repulsing negative pairs from different instances. One of the most important parts in contrastive learning is the selection of positive and negative pairs~\cite{tian2020view} for instance discrimination. To create positive pairs without prior label information, one common way is to generate multiple views of input images. \cite{mocov2, he2020momentum, SimCLR} apply a stochastic augmentation module to randomly augment an input image twice. \cite{LooC} proposes a new model with multi-augmentation and multi-head. It constructs multiple embeddings and captures varying and invariant information introduced by different augmentations. Methods combined contrastive learning with online clustering~\cite{SwAV, Wang_2021_CVPR, li2021prototypical} are proposed to boost the performance of self-supervised learning which explore the data manifold to learn image representations by capturing invariant information.
\section{Method}

\subsection{Contrastive Learning Framework}

We apply MoCo~\cite{he2020momentum} as our unsupervised learning method. MoCo is a mechanism for building dynamic dictionaries for contrastive learning. MoCo defines ``query'' and ``key'', which are representations encoded by encoder networks. Given a batch of inputs, query and positive key are from two different augmentation \emph{views} of the same input whereas query and negative keys are from \emph{views} of different inputs. The dynamic dictionary stores both positive and negative keys. With MoCo, self-supervised learning can be regarded as a training process to perform dictionary look-up. MoCo learns image representations by matching an encoded query $q$ to a dictionary of encoded keys $k$ using a contrastive loss. Query $q$ should be similar to its matching key, positive key $k^+$, and dissimilar to negative keys $k^-$. As illustrated in Figure~\ref{fig:model}, the MoCo model consists of five parts:

A \emph{stochastic data augmentation} module $\mathcal{T}$~\cite{SimCLR} transforms one given input image $I$ into different augmented \emph{views} with randomly applied augmentations, denoted $I_i$ and $I^+$ for query $q_i$ and positive key $k^+$~\cite{he2020momentum}. We sequentially apply five augmentations (Base and Color Aug. in Table~\ref{tab:aug}) similar as MoCo v2~\cite{mocov2}.
    
A \emph{base encoder} $f_q$ for query $q$ maps the augmented \emph{views} into feature space: $f_q: I_i, \rightarrow \boldsymbol{z}_i,\ \text{where}\ I_i\in\mathbb{R}^{C\times W\times H}, \boldsymbol{z}_i\in\mathbb{R}^{s}$, where $z$ denotes the $s$-D encoded \emph{representation}. The parameters are updated by back-propagation. This takes the form of a CNN in our case.
    
A \emph{momentum-updated encoder} $f_k$ for keys $k$ shares the same structure with \emph{base encoder} $f_q$ and is initialized with the same parameters. However, they are not learned through backpropagation during training; instead, the parameters of $f_k(\cdot)$ are updated with a momentum mechanism~\cite{he2020momentum}.
    
A \emph{projection head} projects the \emph{representations} $\boldsymbol{z}_i$ into a unit hypersphere: $h: \boldsymbol{z}_i\rightarrow \boldsymbol{q}_i,\ \text{where}\ \boldsymbol{q}_i\in\mathbb{R}^{m}\ \text{and}\ \left|\left|\boldsymbol{q}_i\right|\right|=1$, the same for $\boldsymbol{z}^+$. The similarity is measured by a dot product.
    
A \emph{dynamic dictionary} holds the prototypical feature for all instances~\cite{Dahua_2018_CVPR, SimCLR, Wang_2021_CVPR}.  It is implemented as a queue of fixed size, fed with the stream of mini-batches that are used for training: in current mini-batch, the encoded representations are enqueued, and the oldest are dequeued~\cite{he2020momentum}.

A popular choice of contrastive loss for positive pairs $(\boldsymbol{q}, \boldsymbol{k}^+)$ and negative pairs $(\boldsymbol{q}, \boldsymbol{k}^-)$ is InfoNCE~\cite{cpc}, denoted as $\mathcal{L}_q(\boldsymbol{q}, \boldsymbol{k}^+)$
\begin{equation}
    \mathcal{L}_q(\boldsymbol{q}, \boldsymbol{k}^+)=-\log\frac{\exp\left(\boldsymbol{q}\!\cdot\!\boldsymbol{k}^+\!/\tau\right)}{\exp\left(\boldsymbol{q}\!\cdot\!\boldsymbol{k}^+\!/\tau\right)+\sum_{k^-} \exp\left(\boldsymbol{q}\!\cdot\! \boldsymbol{k}^-\!/\tau\right)}
\label{eq:moco}
\end{equation}
where the dictionary contains $K$ negative samples and $\tau$ denotes the temperature parameter, which is the hyper-parameter scaling the distribution of distances~\cite{cpc, he2020momentum}.

\subsection{Clustering Based Contrastive Learning}
\begin{figure}[t]
\begin{center}
  \includegraphics[width=\linewidth]{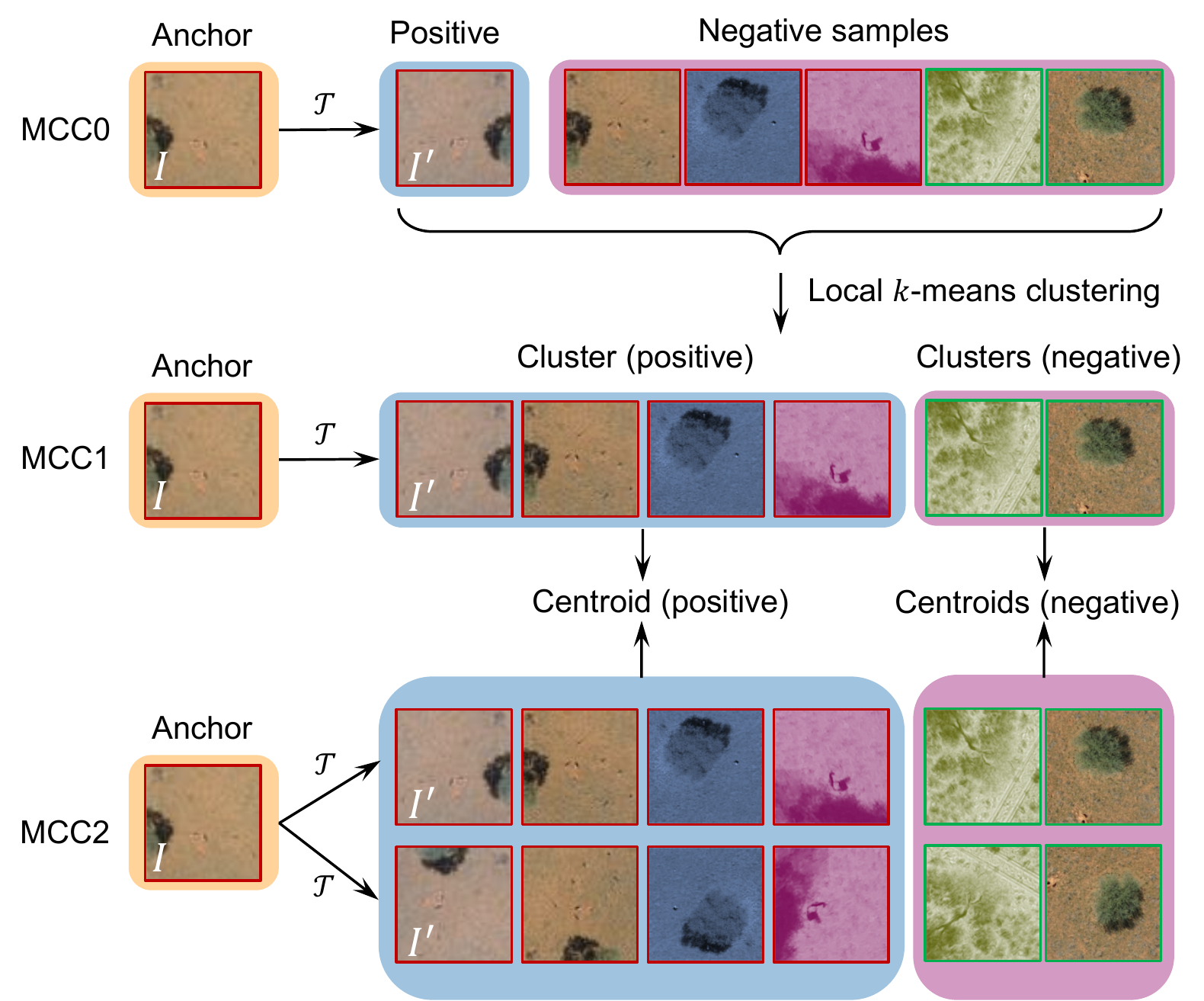}
\end{center}
  \caption{Illustration of positive and negative samples in MCC0, MCC1, and MCC2 scenarioes. Anchor $I$ is the original image patch on KWD dataset. $I^\prime$ is the augmented view of $I$. The \textcolor{red}{red}/\textcolor{green}{green} bounding boxes represent the foreground/background classes. In MCC0, negative samples might contain images from the same category of the anchor (positive samples). That causes the false repulsion. In MCC1 and MCC2, after applying local $k$-means clustering to all samples, positive and negative samples can be grouped into different categories. With CLD, there is less repulsion between samples in the same category.}
\label{fig:positive}
\end{figure}

The key idea of CLD~\cite{Wang_2021_CVPR} is to cluster instances locally, and perform contrastive loss to centroids and image representaions. Therefore, similar instances are clustered into the same group and the false rejection of instances with high similarity is alleviated, as illustrated in Figure~\ref{fig:positive}.  CLD uses two different \emph{views} of the same instance as input. As such, the CLD branch shares the same query encoder $f_q$ with MoCo but uses a different projection head $h_G$, as illustrated in the lower part of Figure~\ref{fig:model}. 

To perform CLD, the unit-length features $\boldsymbol{g}_i$ of all instances in a mini-batch are first extracted from $f_q$ and $h_G$. Then, CLD implements local $k$-means clustering to $\boldsymbol{g}_i$ for a mini-batch of instances and finds $k$ local cluster centroids $\{\boldsymbol{c}_i^1,\dots,\boldsymbol{c}_i^k\}$ with $\boldsymbol{g}_i$ assigned to $\boldsymbol{C}(\boldsymbol{g}_i)$. The same operation is performed to the other branch $I_j$ from all instances in a mini-batch, denoted as $\boldsymbol{g}_j$, $\{\boldsymbol{c}_j^1,\dots,\boldsymbol{c}_j^k\}$, and $\boldsymbol{C}(\boldsymbol{g}_j)$. CLD applies the contrastive loss between $\boldsymbol{g}_i$ and clustering of the other branch $\{\boldsymbol{c}_j^1,\dots,\boldsymbol{c}_j^k\}$. Each cluster contains highly similar instances, and the assigned centroids together with representations from the other branch can be regarded as positive pairs. Namely, the centroids of the \emph{other} clusters act as negative samples~\cite{Wang_2021_CVPR}. Thus, feature vector $\boldsymbol{g}_i$ and its counterparts $\boldsymbol{g}_j$ assigned centroid $\boldsymbol{C}(\boldsymbol{g}_j)$ comprise positive pairs and all \emph{other} centroids comprise negative pairs. The local contrastive loss for CLD is

\begin{equation}
    \mathcal{L}_g(\boldsymbol{g}_i, \boldsymbol{C}(\boldsymbol{g}_j))=-\log\frac{\exp\left(\boldsymbol{g}_i\!\cdot\!\boldsymbol{C}(\boldsymbol{g}_j)\!/\tau\right)}{\sum_{\left\{\boldsymbol{c}_j^k\right\}} \exp\left(\boldsymbol{g}_i\!\cdot\! \boldsymbol{c}_j^k\!/\tau\right)}
\label{eq:cld}
\end{equation} where $\left\{\boldsymbol{c}_j^k\right\}$ denotes the set of $k$ centroids from the other branch. Thus, the loss of a dual-branch CLD in Figure~\ref{fig:model} is:

\begin{equation}
    \mathcal{L}_g(\boldsymbol{g}_1, \boldsymbol{C}(\boldsymbol{g}_2)) +  \mathcal{L}_g(\boldsymbol{g}_2, \boldsymbol{C}(\boldsymbol{g}_1))
\end{equation}

\subsection{Augmentation Strategies}
\label{sec:aug}
\begin{table}
\begin{center}
\begin{tabular}{ll}
\specialrule{.1em}{.05em}{.05em} 
Module & PyTorch-like Augmentation \\
\hline
Base Aug. & RandomCrop(224)$^*$\\
& RandomHorizontalFlip(p=0.5)  \\
& GaussianBlur([0.1, 2.0])\\
\hline
Color Aug. & ColorJitter(0.4, 0.4, 0.4, 0.1)\\
&RandomGrayscale(p=0.2)\\
\hline
Rot. Aug. & RandomRotation()$^{**}$\\
\specialrule{.1em}{.05em}{.05em}
\end{tabular}
\end{center}
\caption{Overview of the employed random augmentation strategies. $^*$ To avoid information loss on tiny animals, we apply random crops without resizing. $^{**}$ We randomly rotate the images by $\{90^\circ$, $180^\circ$, $270^\circ\}$.}
\label{tab:aug}
\end{table}

State-of-the-art contrastive learning~\cite{SimCLR, simsiam, mocov2} between multiple \emph{views} of the data employs stronger augmentation strategies to improve performance. The choices of different \emph{views} have a marked impact on the performance of self-supervised pretraining~\cite{mocov2, what, LooC, tian2020view}. For different branches in CLD, \emph{e.g.}, $I_1$ and $I_2$ in Figure~\ref{fig:model}, we apply multiple augmentations to the same input image~\cite{LooC}. We keep each branch invariant to one specific augmentation transformation. For example, $I_1$ and $I^+$ are always augmented by the same color but different rotation augmentation while $I_2$ and $I^+$ are always augmented by same rotation but different color augmentation. Augmentation parameters are sampled randomly and independently from the stochastic augmentation module $\mathcal{T}$ as outlined in Table~\ref{tab:aug}. We project the queries and key into one embedding space, keeping the embedding space invariant to all augmentations. We aim to add extra geometric transformations on top of the CLD framework. The loss of our proposed augmentation strategies has the same form of Equation~(\ref{eq:moco}) and (\ref{eq:cld}). 

\subsection{Total contrastive loss}
We combine CLD~\cite{Wang_2021_CVPR} with MoCo v2~\cite{mocov2} and construct a total contrastive loss over \emph{views} $I_1$, $I_2$, and $I^+$ with CLD weight $\lambda$ in a mini-batch. We apply different temperatures $\tau_q$ and $\tau_g$ for instance and group branches, respectively. The total contrastive loss is~\cite{Wang_2021_CVPR}:
\begin{equation}
\begin{aligned}
    \mathcal{L}_{tot} = &\frac{1}{2} \left[\mathcal{L}_q(\boldsymbol{q}_1, \boldsymbol{k}^+) + \mathcal{L}_q(\boldsymbol{q}_2, \boldsymbol{k}^+)\right]\\
    &+ \lambda\times\frac{1}{2}\left[\mathcal{L}_g(\boldsymbol{g}_1, \boldsymbol{C}(\boldsymbol{g}_2)) +  \mathcal{L}_g(\boldsymbol{g}_2, \boldsymbol{C}(\boldsymbol{g}_1)\right]
\end{aligned}
\end{equation} where $\boldsymbol{q}_{\{1,2\}}$ and $\boldsymbol{g}_{\{1,2\}}$ are feature representations issued from augmented versions of the original samples, generated following the procedure described in Section~\ref{sec:aug}.


\section{Experiments}

\begin{table*}
\begin{center}
\begin{tabular}{ll|c|c|c|c|c|c}
\specialrule{.1em}{.05em}{.05em} 
&&\multicolumn{3}{c|}{pretraining} & & & Augmentation \\
\cline{3-5}
Name & Backbone & Supervised & Unsupervised & Dataset & Fine-tuning & CLD & Strategies \\
\hline
Sup1 & ResNet-50 & \checkmark &  & ImageNet & &  \\
Sup2 & ResNet-50 & \checkmark &  & ImageNet & \checkmark &  \\
MCC0 & MoCo v2 &  & \checkmark & KWD-Pre & &  \\
MCC1 & MoCo v2  &   & \checkmark & KWD-Pre & & \checkmark \\
MCC2 & MoCo v2  &  & \checkmark & KWD-Pre & & \checkmark & \checkmark\\
\specialrule{.1em}{.05em}{.05em}
\end{tabular}
\end{center}
\caption{Overview of models we use.}
\label{table: models}
\end{table*}

\subsection{Study Area and Data}

In this work, we use the data from~\cite{Beni_2018_RSE}, consisting of RGB aerial images acquired with a SenseFly eBee\footnote{\url{https://www.sensefly.com}} UAV over the Kuzikus Wildlife Reserve in Namibia\footnote{\url{https://kuzikus-namibia.de}} by the SAVMAP consortium\footnote{\url{http://lasig.epfl.ch/savmap}}. The UAV's flight height varied between 120 and 160m, resulting in a resolution of 4 to 8 cm with the given camera (Canon PowerShot S110). The images were annotated with bounding boxes for animals in a crowdsourcing campaign led by MicroMappers\footnote{\url{https://micromappers.wordpress.com}}; these annotations were then refined in several iterations by the authors. This resulted in a total of 1183 animals. We derived the \textbf{K}uzikus \textbf{W}ildlife \textbf{D}ataset \textbf{Pre}-training (KWD-Pre) and \textbf{K}uzikus \textbf{W}ildlife \textbf{D}ataset \textbf{L}ong-\textbf{T}ail distributed (KWD-LT) for pre-traning and fine-tuning/downstream task.

\textbf{Technical Challenges.}
Most contrastive models are trained on curated dataset with unique characteristics, \emph{e.g.}, ImageNet~\cite{imagenet_cvpr09}. In these datasets, images contain only a single object which is located in the center of the image (\emph{object-centric}). And objects have \emph{discriminative} visual features. The datasets also have uniformly distributed classes. In contrast, domain-specific datasets (\emph{e.g.}, our KWD) contain less discriminative visual feature, making it hard to distinguish between similar and small objects \emph{e.g.}, small trees, wildlife from the top view.

\textbf{KWD-Pre.}
We apply the same patches creating procedure as described in~\cite{Beni_2018_RSE}. We randomly crop 15 patches for every original $4000\times3000$ image. The size of each patch is $256\times256$ pixels to save memory and have a larger batch size. We randomly crop 15 extra patches if one image contains animals. Cropping this way increases the chances of extracting patches containing animals for training, but we do not retain any labeled information nor bounding box location. As this can be seen as a form of weak supervision in the patch extraction process, we do not know whether each patch contains animal(s) or not while applying random cropping. So the prior knowledge of classes and locations is not exploited by self-supervised learning. 

\textbf{KWD-LT.}
The original images are taken by UAVs on different dates and times~\cite{Beni_2018_RSE}. We first split the original data into train, test, and validation set with a ratio of 8:1:1. Then, for the background class, we apply a random cropping procedure ($512\times512$ pixels) to the original images and verify each patch to make sure it contains no animal. For the foreground (wildlife) class, we apply a random cropping procedure ($224\times224$ pixels) around the ground truth bounding boxes to make sure each patch contains whole animal(s) body. We choose three different random seed to random cropping procedures of train, test, and validation set to make sure the cropping position is different. The train set is class imbalanced and long-tail distributed with a foreground-to-background ratio of $\frac{1}{18}$. The test and validation set are class balanced. In the experiments below, we evaluate fine-tuning on KWD-LT with different percentages of annotated animals to investigate the benefit of SSL for reduced annotation efforts.



\begin{figure}[t]
\begin{center}
  \includegraphics[width=0.86\linewidth]{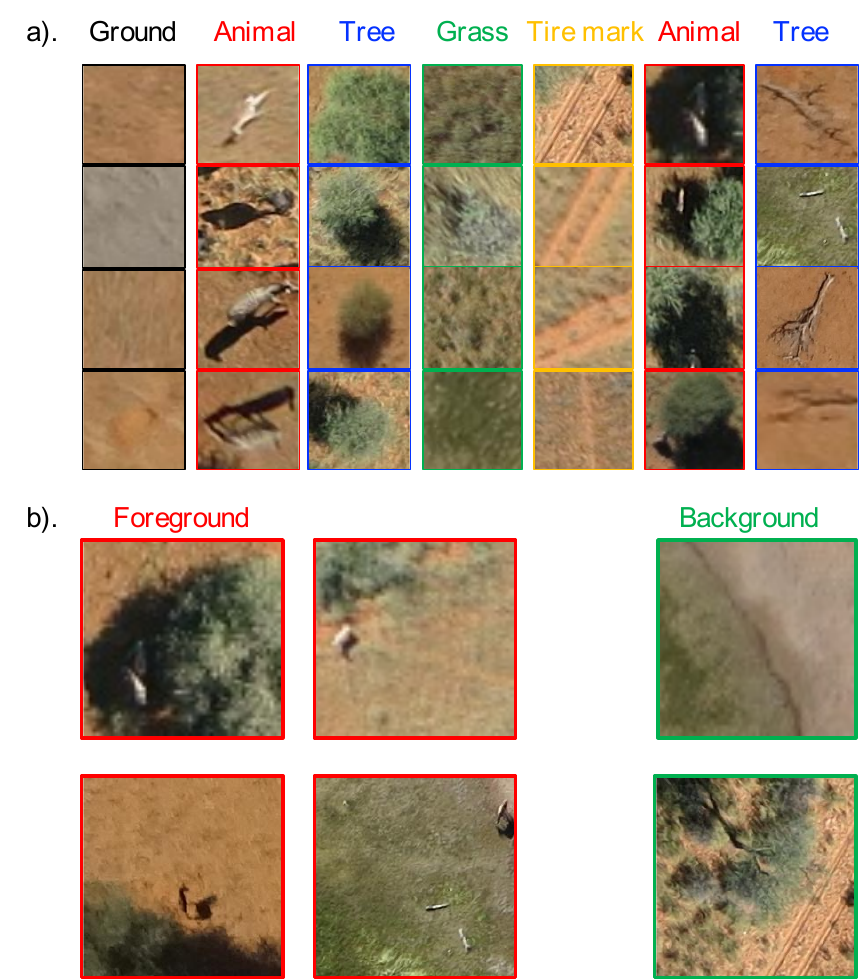}
\end{center}
  \caption{Overview of KWD-LT dataset. (a) denotes all possible elements in the dataset. Especially the animal elements in the right are animals beneath the tree and the tree elements in the right are dead tree trunks. This is extremely hard to be recognized~\cite{Beni_2018_RSE}. It is also difficult to distinguish between dead tree trunks and animal. Intuitively, all examples in the KWD-LT dataset consist of randomly combined elements in (a). (b) is examples of KWD-LT dataset. Instances are image-level annotated. Foreground represent images containing wildlife.}
\label{fig:dataset}
\end{figure}

\subsection{Experimental Setup}

\textbf{Models.}
For the supervised models, we use a ResNet-50~\cite{resnet}, pretrained on ImageNet~\cite{imagenet_cvpr09}. Sup1 freezes the output of ResNet-50 average pooling layer. Sup2 fine-tunes the pretrained ResNet-50 on KWD-LT with full labels, as shown in Figure~\ref{table: models}. We apply the MoCo v2 as our contrastive baseline model, denoted as MCC0. Instead of using a RandomResizeCrop, we apply a PyTorch RondomCrop to input images to keep more information. MCC1 and MCC2 are all MoCo v2 model with CLD. We apply our augmentation strategies to MCC2. We set the $\lambda$, number of clusters to 0.25 and 32 respectively. Detailed information of different models are outlined in Table~\ref{table: models}.

\textbf{Optimizer.}
We use stochastic gradient descent for self-supervised pretraining and downstream task fine-tuning. For the self-supervised pretraining, we apply the same cosine decay scheduler as proposed in~\cite{Wang_2021_CVPR}. For the semi-supervised fine-tuning, the initial learning rate is 0.01 and we likewise apply a cosine decay schedule~\cite{SimCLR}. For the downstream task, we set the initial learning rate as 30 and we apply same strategy proposed in~\cite{he2020momentum}. 


\textbf{Base Encoder, Projection Head.} 
We apply a ResNet-50 without pretrained on ImageNet as our base encoder. We simply remove the last fully-connected layer and use the output of average pooling as feature vector $\boldsymbol{z}_i$. We adopt a Multi-Layer Perceptron (MLP) head following~\cite{SimCLR, mocov2}, which is a 2-layer MLP (2048-dimensional hidden layer, with ReLU). We share the hidden layer and apply a different final layer of the MLP head to MoCo branch and CLD branch. The dimension of the unit-length feature representation $q_i$ and $k^{\{+,-\}}$ is 128.

\textbf{Hyperparameter Choice.}
For fair comparison and avoiding hyper-parameters tuning redundancy, we select the CLD weight $\lambda$ and number of cluster by linear classification on frozen features with labeled data. According to our prior knowledge, the images in the KWD dataset are primarily composed of artefacts as follows: animal, tree, grass, tire mark (road), animal beneath tree, and dead tree trunk, as illustrated in Figure~\ref{fig:dataset}. Ideally, the number of all possible clusters is therefore $C_6^0 + C_6^1 + C_6^2 + C_6^3 + C_6^4 + C_6^5 + C_6^6 = 2^6 = 64$. Among all elements, animals beneath tree are hard to recognize; also dead tree trunks are easy confused with animals. Meanwhile, with a batch size of 64, it might not yield all 64 combinations, but we still use 64 clusters to give the model enough freedom. We train the MCC1 model with different hyper-parameters for 200 epochs and select the best hyper-parameter combination based on accuracy on the validation set of the downstream recognition task.


\textbf{Downstream Recognition Task.}
We verify different models by applying linear classification on encoded image representations. We follow the same common linear classification protocol as~\cite{he2020momentum}. We first perform self-supervised pretraining on KWD-Pre dataset. Then we perform two kinds of experiments: (1) \emph{frozen} features: we freeze the output features of the global average pooling layer of a ResNet and train a linear classifier (a fully-connected layer followed by softmax)~\cite{he2020momentum} in a supervised way on our KWD-LT downstream task dataset; (2) \emph{end-to-end}: we fine-tune the base encoder and linear classifier by softmax loss instead of contrastive loss. And we report the linear classification top-1 accuracy on the KWD-LT validation set, as well as recall and precision for foreground class.

\section{Results and Discussion}

\subsection{Hyperparameter Choice}

The results for the ablation studies on the number of clusters for CLD and $\lambda$ are shown in Tables~\ref{tab:lambda} and~\ref{tab:cluster}. We can clearly see that the model performs almost equally when $\lambda = 0.25$ or $\lambda = 1$. Hence, we chose the default $\lambda = 0.25$ value as in~\cite{Wang_2021_CVPR}. However for the number of clusters, there does not seem to be an obvious trend. we speculate that (1) grouping projected \emph{representations} by $k$-means clustering is hard to perform well, (2) the model is not able to recognize animals beneath the tree or/and to distinguish between dead tree trunks and animals (number of clusters = 16 or 32). When number of clusters = 16, 32, or 64, the accuracy is equally good. Therefore, we chose the best recall with number of cluster = 32.

\begin{table}
\begin{center}
\begin{tabular}{lcccc}
\specialrule{.1em}{.05em}{.05em} 
Model & $\lambda$ & Acc & Prec & Rec \\
\hline
MCC1 & 0.1 & 86.2 & 98.6 & 73.4\\
MCC1 & \textbf{0.25} & \textbf{88.4} & 98.9 & \textbf{77.4}\\
MCC1 & 0.5 & 86.4 & 98.9 & 73.5 \\
MCC1 & 1 & 88.1 & 98.5 & 77.2\\
CLD & 1 & 60.9 & 98.9 & 22.1\\
\specialrule{.1em}{.05em}{.05em} 
\end{tabular}
\end{center}
\caption{Hyper-parameter $\lambda$ selection. All models are pretrained on KWD-Pre using the MCC1 strategy.}
\label{tab:lambda}
\end{table}

\begin{table}
\begin{center}
\begin{tabular}{lcccc}
\specialrule{.1em}{.05em}{.05em} 
Model & Clusters & Acc & Prec & Rec \\
\hline
MCC1 & \textbf{16} & 88.3 & 100.0 & 76.5 \\
MCC1 & 30 & 87.3 & 99.7 & 75.0 \\
MCC1 & \textbf{32} & \textbf{88.4} & 98.9 & \textbf{77.4}\\
MCC1 & 48 & 87.2 & 99.7 & 74.9\\
MCC1 & \textbf{64} & \textbf{88.4} & 100.0 & 76.9\\
\specialrule{.1em}{.05em}{.05em} 
\end{tabular}
\end{center}
\caption{Number of clusters selection. All models are pretrained on KWD-Pre using the MCC1 strategy with $\lambda = 0.25$.}
\label{tab:cluster}
\end{table}

\subsection{Main Results}

\begin{table}
\centering
\begin{tabular}{lcccc}
\specialrule{.1em}{.05em}{.05em}
Model & Epochs$^*$ & Acc & Prec & Rec \\
\hline
Sup1 & - & 86.7 & 96.5 & 76.1 \\
Sup2 & 200 & 88.6 & 99.3 & 77.7 \\
\hline
MCC0 & 200 & 82.2 & 97.7 & 65.6\\
MCC1 & 200 & 88.4 & 98.9 & 77.4 \\
MCC2 & 150 & 90.8 & 99.6 & 81.9 \\
\specialrule{.1em}{.05em}{.05em} 
\end{tabular}
\caption{\textbf{Linear classifier} top-1 accuracy (\%), foreground class precision and recall (\%) on \emph{frozen} features with full labels, comparison of self-supervised learning on KWD-Pre (MCC0, MCC1, MCC2) and supervised pretraining on ImageNet (Sup1, Sup2). $^*$ We adopt the peak performance epoch to compare different methods.}
\label{tab:linear}
\end{table}

\begin{figure}[t]
\begin{center}
  \includegraphics[width=\linewidth]{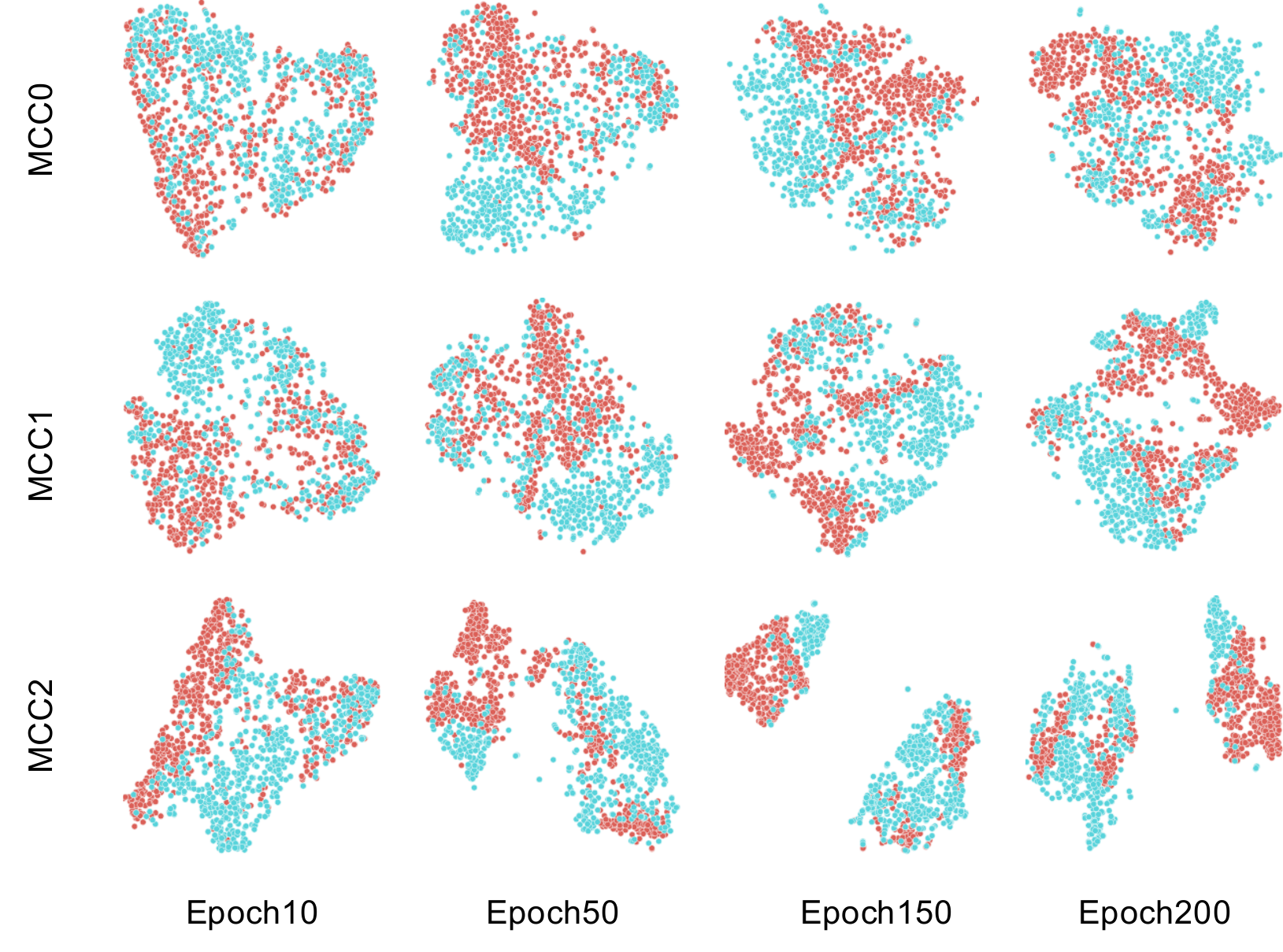}
\end{center}
  \caption{\textbf{t-SNE feature visualization} of MCC0, MCC1, and MCC2 on KWD-LT. MCC2 has earlier and better separation between foreground and background classes (indicated by color) than MCC0 and MCC1.}
\label{fig:tsne}
\end{figure}

\begin{table*}
\begin{center}
\setlength{\tabcolsep}{4pt}
\begin{tabular}{lll|ccc|ccc|ccc}
\specialrule{.1em}{.05em}{.05em} 
Pretraining & SSL &Fine-tuning&\multicolumn{3}{c|}{1\% labels}& \multicolumn{3}{c|}{10\% labels}  & \multicolumn{3}{c}{20\% labels} \\
dataset& strategy & (on KWD-LT)& Acc & Prec &Rec & Acc & Prec & Rec & Acc & Prec & Rec\\
\hline
\multirow{2}{*}{ImageNet}&\multirow{2}{*}{-}&\emph{end-to-end}& 50.4 & 0 & 0 & 69.8 & 98.8 & 40.2 & 80.5 & 97.3 & 62.8\\
&&\emph{frozen} features & 54.4 & 100.0 & 9.0 & 69.6 & 99.3 & 39.3 & 76.2 & 99.5 & 52.7\\
\hline
\multirow{2}{*}{KWD-Pre}&\multirow{2}{*}{MCC0}&\emph{end-to-end} & 68.0 & 98.5 & 35.9 & 76.8 & 99.7 & 53.4 & 77.9 & 99.7 & 55.4 \\
&&\emph{frozen} features & 74.0 & 97.1 & 49.1 & 76.9 & 96.8 & 55.1 & 77.4 & 97.3 & 56.1\\\hline
\multirow{2}{*}{KWD-Pre}&\multirow{2}{*}{MCC1}&\emph{end-to-end} & 70.5 & 98.5 & 40.9 & 76.0 & 99.7 & 51.6 & 88.5 & 99.3 & 77.4 \\
&&\emph{frozen} features & 71.9 & 94.8 & 46.0 & 82.5 & 98.2 & 66.0 & 83.8 & 98.2 & 68.8\\\hline
\multirow{2}{*}{KWD-Pre}&\multirow{2}{*}{MCC2}&\emph{end-to-end} & 78.9 & 98.0 & 58.7 & 90.7 & 99.8 & 81.3 &\textbf{91.9}&100&83.7\\
&&\emph{frozen} features & \textbf{83.4}&97.0&68.6& \textbf{91.7}&98.8&84.5&90.1&99.1&81.4\\
\specialrule{.1em}{.05em}{.05em} 
\end{tabular}
\end{center}
\caption{\textbf{Animal recognition} accuracy when using a portion of the available labeled samples in the final classifier (\emph{frozen} features) or in the base encoder (\emph{end-to-end}) and when varying the type of pretraining, the self-supervised strategy and the type of fine tuning. Prec = Precision, Rec = Recall.}
\label{tab:fewshot}
\end{table*}

\textbf{Results of self-supervised pretraining.}
The possible repeated and highly correlated patches slow the training process and lower the performance of SSL pretraining. As it breaks the instance discrimination presumption described in Section~\ref{intro}. For fair comparison, we evaluate the performance of self-supervised pretraining by linear classification of \emph{frozen} features with full labels. The linear classifier accuracy in Table~\ref{tab:linear} shows that SSL pretraining on target dataset is a strong competition of supervised pretraining on ImageNet. MoCo with CLD (MCC1) perform equally as well as supervised fine-tuning (Sup2). With controlled and designed augmentation for our scenario, MCC2 outperforms Sup2 by 2.2\%. MCC2 at epoch 150 outperforms MCC0 at epoch 200 by 8.6\% (82.2\% vs. 90.8\%) with a faster converging. We can tell that SSL pretraining with controlled augmentation can improve the performance of rare wildlife recogniztion. However, the accuracy of MCC2 in epoch 200 is lower than that in epoch 150. That might imply that adding geometric transformation might cause the problem of overfitting. And feature visualization in Figure~\ref{fig:tsne} show that CLD with geometric augmentation (MCC2) converges faster and better towards a more distinctive feature representation than MCC0 and MCC1.

\textbf{Results of Recognition Task.} 
Self-supervised pretraining on our dataset can utilize annotations far more efficiently than supervised pretraining on ImageNet. As shown in Table~\ref{tab:fewshot}, fine-tuning the encoder \emph{end-to-end} will completely destroy the capacity of ImageNet pretrained model. Even though we freeze the feature representations, the model perform poorly with small fraction of labeled instances. However, fine-tuning only linear classifier on \emph{frozen} features with 1\% and 10\% annotations outperforms fine-tuning the encoder \emph{end-to-end}. When we only train the linear classifier with 1\% of labels, MCC2 outperforms MCC0 by 9.4\% and MCC1 by 11.5\%. For MCC1, fine-tuning encoder with 10\% annotations outperforms the recognition accuracy with full annotations. For MCC2, we need 20\% annotations to get a better result. Meanwhile, for MCC1, fine-tuning the encoder with 20\% of labeled instances will have the same performance with fine-tuning ImageNet pretrained model (Sup2) with full labeled instances. Whereas for MCC2, only 10\% labeled data is required to outperform the Sup2 supervised model. The results show that SSL model can learn more information through geometric invariant mapping and capturing geometric invariant information can benefit UAV top view imagery task. 

\textbf{Influence of Label Fraction.}
The results of Sup1, Sup2, MCC0, and MCC1 in Table~\ref{tab:linear} and of these \emph{frozen} features in Table~\ref{tab:fewshot} show that increasing the fraction of labels used can improve the performance of linear classification. But it has a bottleneck performance of 88.6\%. This is different in training linear classifier on MCC2. Although adding extra geometric augmentation can boost performance. Feeding all annotated data can decrease the performance of \emph{frozen} features in MCC2, as shown in Table~\ref{tab:linear} and~\ref{tab:fewshot}. That could be caused by imbalanced classes: the model overfits to the background class and is over-confident. 
\section{Conclusion}
In this paper, our proposed strategy reduces the label requirements in the wildlife recognition tasks. The problem is introduced by applying supervised learning to automated animal censuses in largely remote areas with aerial imagery, in which scenario the annotations are expensive to be obtained. The contrastive self-supervised pretraining with domain-specific geometric transformation outperforms the performance of fine-tuning ImageNet pretrained model with full labels. Results show that the geometric invariant mapping method can capture information more efficiently of wildlife in UAV images than method without geometric augmentation. Extensive experiments further prove the effectiveness of recognizing rare wildlife with reduced labels.

\ 

\textbf{Acknowledgments.}
The authors would like to thank the Kuzikus Wildlife Reserve, Namibia\footnote{\url{https://www.kuzikus-namibia.de}} for the access to the aerial data and the ground reference used in this study and Mei Sun of IfU, ETH Zürich, Xia Li of CVL, ETH Zürich for their helpful discussion. 

{\small
\bibliographystyle{ieee_fullname}
\bibliography{ICCV_LUAI_12_CR}
}

\end{document}